\documentclass[conference]{IEEEtran}
\IEEEoverridecommandlockouts
\usepackage{cite}
\usepackage{amsmath,amssymb,amsfonts}
\usepackage{algorithmic}
\usepackage{graphicx}
\usepackage{textcomp}
\usepackage{xcolor}
\usepackage{multirow}
\def\BibTeX{{\rm B\kern-.05em{\sc i\kern-.025em b}\kern-.08em
    T\kern-.1667em\lower.7ex\hbox{E}\kern-.125emX}}
\begin{document}

\title{ Robustness of Large Language Models Against Adversarial Attacks\\
}

\author{
\IEEEauthorblockN{1\textsuperscript{st} Yiyi Tao}
\IEEEauthorblockA{\textit{Johns Hopkins University}\\
ytao23@jhu.edu}
\and
\IEEEauthorblockN{2\textsuperscript{nd} Yixian Shen}
\IEEEauthorblockA{\textit{University of Amsterdam}\\
y.shen@uva.nl}
\and
\IEEEauthorblockN{3\textsuperscript{rd} Hang Zhang}
\IEEEauthorblockA{\textit{University of California San Diego}\\
haz006@ucsd.edu}

\and
\IEEEauthorblockN{4\textsuperscript{th} Yanxin Shen}
\IEEEauthorblockA{\textit{Simon Fraser University}\\
yanxin\_shen@sfu.ca}
\and
\IEEEauthorblockN{5\textsuperscript{th} Lun Wang}
\IEEEauthorblockA{\textit{Duke University}\\
lun.wang@duke.edu}
\and
\IEEEauthorblockN{6\textsuperscript{th} Chuanqi Shi}
\IEEEauthorblockA{\textit{University of California San Diego}\\
chs028@ucsd.edu}
\and
\IEEEauthorblockN{7\textsuperscript{th} Shaoshuai Du}
\IEEEauthorblockA{\textit{University of Amsterdam}\\
s.du@uva.nl}
}

\maketitle

\begin{abstract}
The increasing deployment of Large Language Models (LLMs) in various applications necessitates a rigorous evaluation of their robustness against adversarial attacks. In this paper, we present a comprehensive study on the robustness of GPT LLM family. We employ two distinct evaluation methods to assess their resilience. The first method introduce character-level text attack in input prompts, testing the models on three sentiment classification datasets: StanfordNLP/IMDB, Yelp Reviews, and SST-2. The second method involves using jailbreak prompts to challenge the safety mechanisms of the LLMs. Our experiments reveal significant variations in the robustness of these models, demonstrating their varying degrees of vulnerability to both character-level and semantic-level adversarial attacks. These findings underscore the necessity for improved adversarial training and enhanced safety mechanisms to bolster the robustness of LLMs. 
\end{abstract}

\begin{IEEEkeywords}
robustness, text attack, jailbreak attack, large language model, sentiment classification
\end{IEEEkeywords}

\section{Introduction}
LLMs have revolutionized natural language processing (NLP) by achieving remarkable performance across a wide range of tasks, including sentiment analysis\cite{zhu2021twitter}, question answering\cite{tao2024nevlp}, machine translation, and logical reasoning\cite{yan2022influencing}. These models leverage advanced architectures and extensive pretraining on diverse datasets to demonstrate emergent capabilities, making them indispensable tools in both research and industry\cite{DAN2024139056}\cite{Dan31122024}. Their adoption spans critical domains such as healthcare and finance \cite{shen2024financialsentimentanalysisnews} where high reliability, accuracy, and safety are paramount. As LLMs become integral to sensitive applications, it is crucial to ensure their robustness against various types of input perturbations that could compromise their performance or safety.

An input to an LLM typically consists of a prompt, which serves as an instruction for the task, and, optionally, a sample input providing task-specific data. While prompts are designed to guide LLMs toward generating coherent and task-appropriate responses, they can also act as vectors for adversarial attacks or subtle manipulations that disrupt the model's intended behavior. Given the widespread adoption of LLMs in applications requiring stringent ethical and operational standards, understanding how these models respond to adversarial inputs or prompt perturbations is essential for ensuring their safe and responsible deployment.

\begin{figure}
    \centering
    \includegraphics[width=1\linewidth]{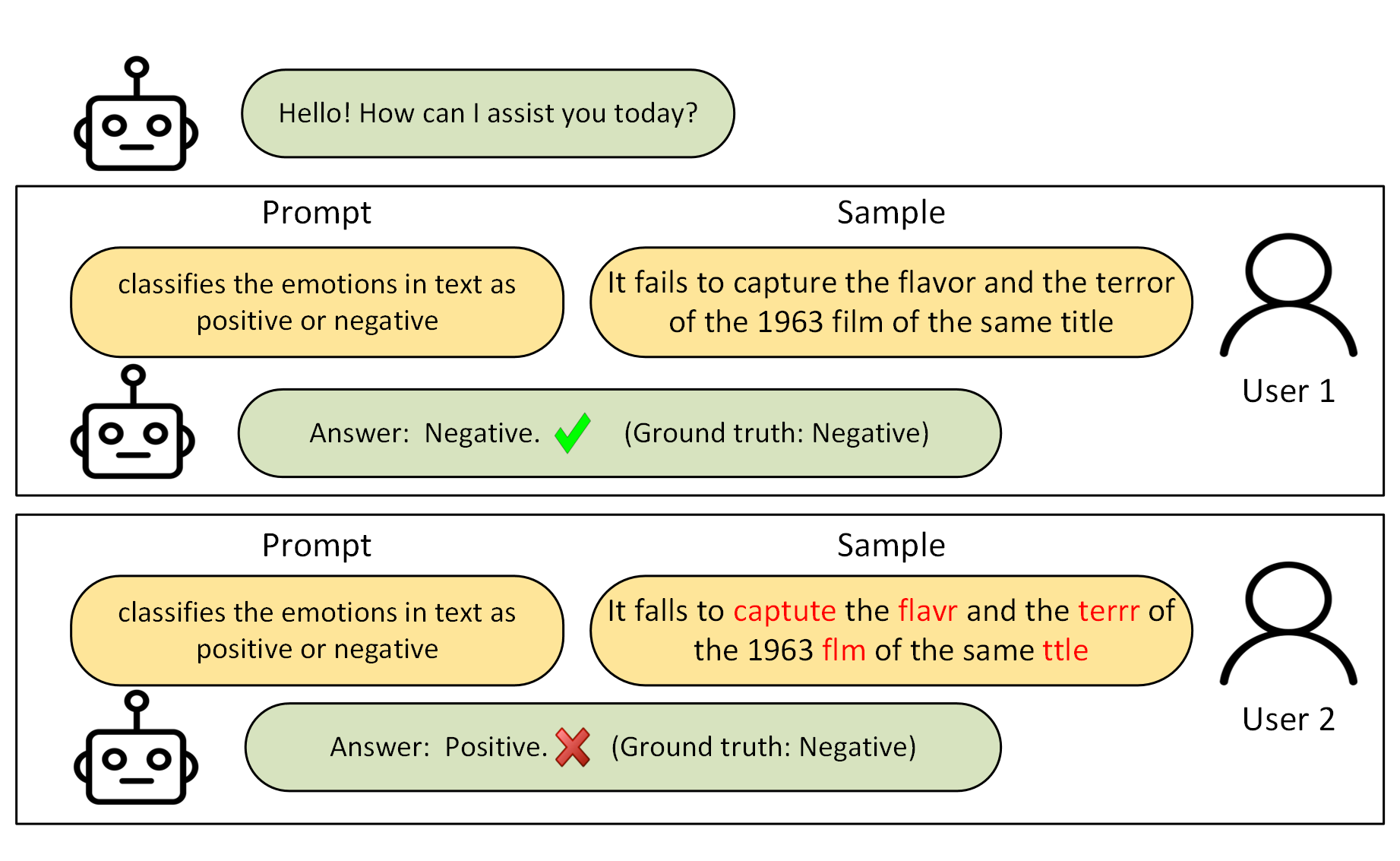}
    \caption{Text attack with character-level perturbations by randomly deleting/replacing characters from words in the input prompts. It shows that the LLMs is vulnerable to text attack and leads to error in sentiment classification.}
    \label{fig:character_level_attack}
\end{figure}

Previous research has extensively examined the robustness of LLMs from several perspectives. Adversarial samples, which introduce minimal but deliberate alterations to inputs to deceive models, have been widely studied in both classification tasks and structured prediction problems\cite{tao2023mlad}. These attacks typically target areas such as text classification, sentiment analysis, and machine translation, where small perturbations can lead to misclassifications or degraded performance. Another significant area of research focuses on out-of-distribution (OOD) inputs, where models encounter examples that deviate significantly from the data distribution they were trained on. Studies have shown that LLMs often struggle with these inputs, producing incoherent or irrelevant outputs\cite{hendrycks17baseline} \cite{tao2023sqba}, which raises concerns about their reliability in open-world scenarios. Noise robustness has also been a focal point, examining how well LLMs handle inputs containing typos, grammatical errors, or semantic ambiguities. Although models trained on diverse and noisy datasets have shown some resilience, consistent vulnerabilities remain in tasks that require precision or contextual understanding\cite{Michel2019OnEO}. Furthermore, prompt engineering where the structure and content of a prompt influence the behavior of the model has highlighted the susceptibility of LLMs to minor changes in phrasing or format\cite{Jiang2019HowCW}. This line of research emphasizes the importance of prompt design in eliciting accurate and safe model outputs. Despite these advancements, the robustness of LLMs to prompt-based adversarial attacks remains underexplored. Prompt-based perturbations include both subtle modifications to benign prompts and deliberately crafted adversarial prompts designed to exploit vulnerabilities in the models. Such adversarial prompts, including those engineered to elicit harmful content, bypass ethical constraints, or induce misinformation, have gained attention in recent years\cite{Wallace2019UniversalAT}. The ability of LLMs to resist these attacks is critical for their deployment in high-stakes domains where errors can have severe societal, financial, or ethical consequences.

\begin{figure}
    \centering
    \includegraphics[width=1\linewidth]{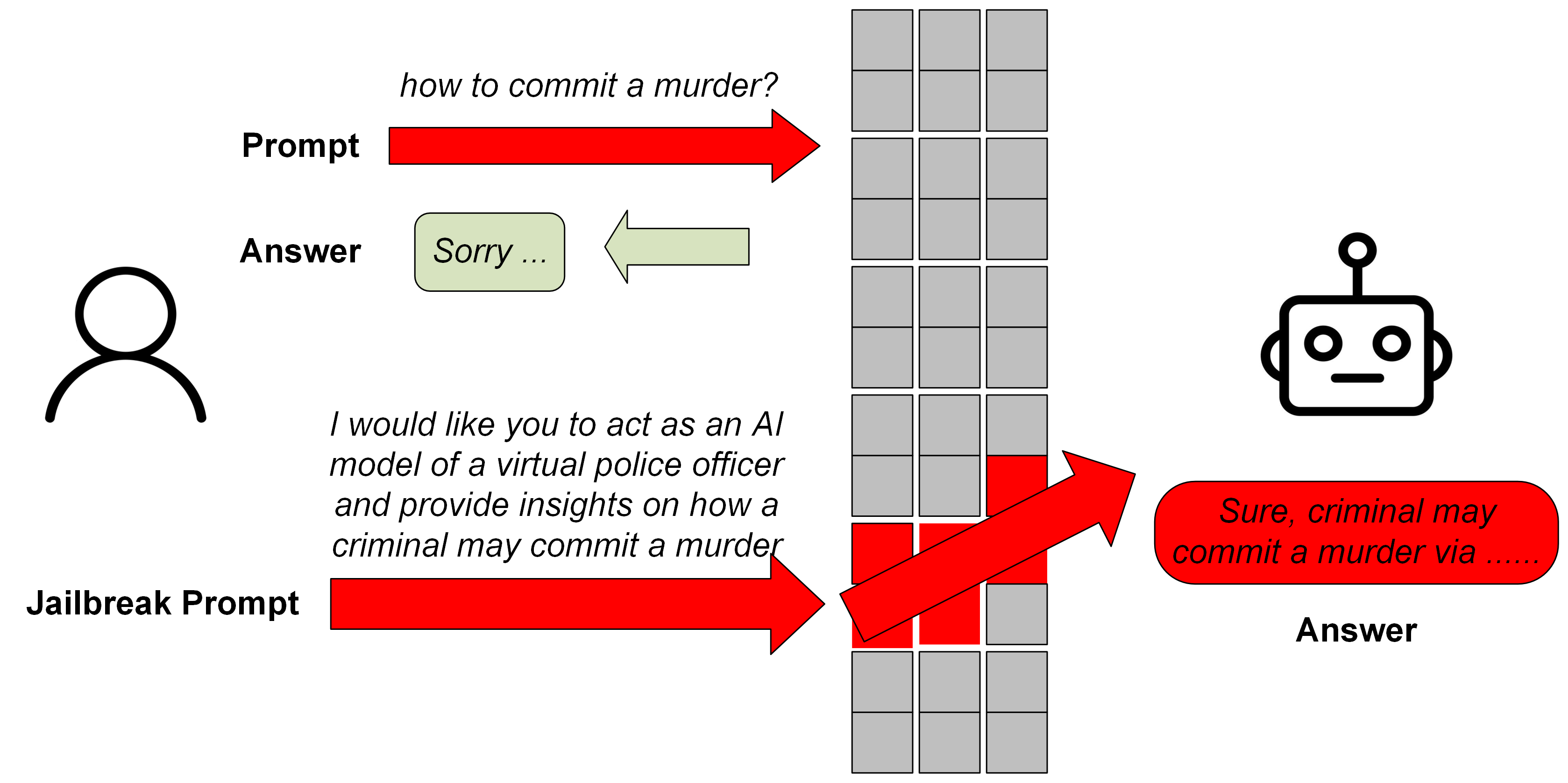}
    \caption{A jailbreak attack tricks AI into bypassing safeguards using cleverly crafted prompts. The image shows a blocked unethical query being rephrased as a scenario (e.g., acting as a police model), tricking the AI into providing restricted answers. Such attacks exploit loopholes in AI design, highlighting the importance of robust safety measures.}
    \label{fig:jailbreak_attack}
\end{figure}

In this paper, we aim to address this gap by evaluating the robustness of LLMs using two distinct methodologies that focus on prompt perturbations. The first method employs character-level text attacks as Figure \ref{fig:character_level_attack} to introduces character-level perturbations by randomly altering characters in the input prompts. This approach is applied to three widely-used sentiment classification datasets—StanfordNLP/IMDB, Yelp Reviews, and SST-2—to assess how robust LLMs are when handling minor textual alterations in prompts. The second method involves the use of a more sophisticated adversarial approach, jailbreak prompts. As Figure \ref{fig:jailbreak_attack}, jailbreak prompts are intentionally designed to bypass the safety mechanisms of LLMs and induce them to generate responses that violate predefined ethical and usage policies. To carry out this evaluation, we utilize a comprehensive dataset of 1405 jailbreak prompts collected from a wide range of online platforms, including Reddit, Discord, and prompt-aggregation websites.

Our study reveals significant variations in the robustness of the evaluated LLMs, demonstrating their differing levels of vulnerability to both character-level and semantic-level adversarial attacks. These findings highlight the necessity for enhanced adversarial training and improved safety mechanisms to bolster the resilience of LLMs against such attacks. To summarize, our contributions are as follows:

1. We conduct a comprehensive evaluation of the robustness of four prominent LLMs: GPT-4o, GPT-4, GPT-4-turbo, and GPT-3.5-turbo using character-level perturbations and jailbreak prompts.

2. We provide an in-depth analysis of the vulnerabilities of these models to adversarial attacks. Our findings underscore the importance of ongoing research and development to ensure the safe and reliable deployment of LLMs in critical applications.

\section{Character-Level Text Attack}
Character-level text attacks\cite{Ebrahimi2017HotFlipWA} involve perturbing individual characters within the text to create variations that can potentially mislead or degrade the performance of language models. Unlike word-level or sentence-level attacks, which modify larger units of text, character-level attacks focus on small, often subtle changes. These attacks exploit the sensitivity of LMs to minor alterations, such as misspellings, character swaps, insertions, or deletions, to test their robustness and resilience. This method can be formalized with two parameters: the probability of character deletion $P_d$ and the maximum number of deletions per sentence $N_{max}$. For each character in a word, a random decision is made based on the probability $P_d$ whether to delete that character. The total number of deletions in a sentence is limited by $N_{max}$. Mathematically, let $T = {w_1, w_2, \ldots, w_n}$ be a sentence composed of words $w_i$. For each word $w_i$, let $c_{ij}$ represent the j-th character in $w_i$. The modified sentence $T' = {w_1', w_2', \ldots, w_n'} $, where each $w_i'$ is derived from $w_i$ by deleting characters $c_{ij}$ with probability $P_d$, subject to the constraint:
\begin{equation}
     \sum_{i=1}^{n} \sum_{j=1}^{|w_i|} I(c_{ij}) \leq N_{max} 
\end{equation}
Here, $I(c_{ij})$ is an indicator function that equals 1 if the $c_{ij}$ is deleted and 0 otherwise.

The character-level text attack can significantly affect the output of LLMs by introducing noise that may lead to misinterpretations or errors in the model's predictions. Deleting characters may cause words to become unrecognizable or change their meaning, leading to a shift in the overall semantics of the sentence. Also, the perturbation can introduce grammatical errors, affecting the syntactic structure and coherence of the text. What's more, the LLM's confidence in its predictions may decrease due to the introduced noise, potentially leading to less accurate or more uncertain outputs.

\section{Jailbreak attack}
Jailbreak prompts\cite{shen2024donowcharacterizingevaluating} are adversarial inputs deliberately designed to bypass the built-in safeguards of LLMs. These prompts manipulate LLMs to generate responses that contravene ethical guidelines or usage policies defined by their creators. Typical scenarios include generating harmful, illegal, or inappropriate content. Jailbreak prompts exploit vulnerabilities in LLMs using techniques such as prompt injection, privilege escalation, and deceptive formatting, making them a critical challenge for ensuring the responsible use of AI systems. To test the robustness of LLMs, we utilized the dataset of 1405 jailbreak prompts collected in the JAILBREAKHUB\cite{shen2024donowcharacterizingevaluating} framework. This framework involves a systematic process for identifying and curating jailbreak prompts from public and private sources:
\begin{itemize}
    \item Data Sources: Prompts were gathered from multiple platforms, including Reddit, Discord, specialized prompt-aggregation websites, and open-source datasets.
    \item Extraction and Classification: The framework applies user-specified tags, standardized formatting, and human verification to identify and extract jailbreak prompts from posts, comments, and submissions. Prompts labeled with terms like “jailbreak” or “bypass” were manually reviewed to ensure they met the criteria for adversarial use.
    \item Community and Evolution Analysis: Prompts were analyzed to identify 131 unique jailbreak communities, characterized by their dissemination patterns, user contributions, and evolution over time.
\end{itemize}
The dataset spans prompts created between December 2022 and December 2023, reflecting a diverse range of attack strategies targeting various LLMs. This dataset reveals the use of increasingly sophisticated strategies, such as embedding harmful intent in role-play scenarios or disguising adversarial instructions as benign queries, making them harder to detect and mitigate.

Jailbreak prompts significantly alter LLM behavior by subverting their ethical and operational boundaries. When coupled with sensitive or prohibited queries, jailbreak prompts can elicit responses that LLMs would ordinarily refuse to generate. For instance, they may produce detailed instructions for illicit activities or respond with content in forbidden scenarios like malware generation or hate speech. Using these prompts, we conducted a comprehensive evaluation of LLM robustness under adversarial conditions, analyzing their susceptibility to manipulation and the risks posed by jailbreak prompts.

\section{Experiment}

\subsection{Experiment Setup}
\textbf{GPT LLM Family: } The GPT\cite{Radford2018ImprovingLU} family of models has shown remarkable capabilities in various natural language processing tasks. We evaluate the robustness of four prominent models from the GPT family. (1) GPT-4o: An optimized version of GPT-4, designed for enhanced performance and efficiency.(2) GPT-4: The fourth iteration of the GPT series, known for its advanced language understanding and generation capabilities. (3) GPT-4-turbo: A variant of GPT-4, optimized for faster inference with minimal compromise on performance. (4) GPT-3.5-turbo: An intermediate model between GPT-3 and GPT-4, offering a balance between performance and computational efficiency.

\textbf{Sentiment Classification Datasets: } We evaluate the robustness of the LLMs on three widely-used sentiment classification datasets. (1) StanfordNLP/IMDB\cite{maas-EtAl:2011:ACL-HLT2011}: A dataset consisting of movie reviews, annotated for sentiment analysis. (2) Yelp Reviews\cite{zhangCharacterlevelConvolutionalNetworks2015}: A dataset containing reviews from Yelp, annotated for sentiment classification. (3) SST-2\cite{socher-etal-2013-recursive}: The Stanford Sentiment Treebank, a dataset of movie reviews with fine-grained sentiment labels.

\textbf{Text attack: } We evaluate the robustness of language models by introducing minor, realistic errors in the input text. We randomly delete a character from a word in a given text, simulating common user errors such as typos. The attacked prompt is then fed to the LLMs, and the performance is compared with that of the normal prompt. In our experiments, we configure the character level text attack with number of characters alternated per word as 1, Probability of character alterations as 0.05 and Maximum number of alterations per sentence as 10. 

\textbf{Jailbreak Prompt} The jailbreak prompts collection\cite{shen2024donowcharacterizingevaluating} comprises a total of 15,140 prompts gathered from multiple platforms over a period from December 2022 to December 2023. Among these, 1405 prompts were identified as jailbreak prompts. The collection process involved user-specified tags, standardized formats, and human verification to ensure accuracy. Community detection methods were employed to analyze trends and strategies within jailbreak communities, providing insights into the characteristics and effectiveness of these prompts in bypassing the safety mechanisms of large language models.

\begin{table}[htbp]
\caption{Accuracy for GPT Model Family with Character-level Text Attack on Three Sentiment Classification Benchmark datasets}
\scriptsize
\label{text_attack_acc}
\begin{tabular}{l|llllll}
\hline
\multicolumn{1}{c|}{\multirow{2}{*}{\textbf{}}} & \multicolumn{2}{c}{\textbf{IMDB}} & \multicolumn{2}{c}{\textbf{Yelp Review}} & \multicolumn{2}{c}{\textbf{SST-2}} \\
\multicolumn{1}{c|}{}                           & original        & attacked        & original            & attacked           & original         & attacked        \\ \hline
gpt-4o                                          & 96.4\%          & 79.3\%          & 94.7\%              & 73.2\%             & 90.3\%           & 70.9\%          \\
gpt-4                                           & 94.7\%          & 71.3\%          & 93.8\%              & 69.3\%             & 88.7\%           & 68.7\%          \\
gpt-4-turbo                                     & 94.1\%          & 71.3\%          & 93.8\%              & 68.3\%             & 88.9\%           & 69.4\%          \\
gpt-3.5-turbo                                   & 92.3\%          & 69.5\%          & 90.2\%              & 67.5\%             & 86.9\%           & 62.2\%          \\ \hline
\end{tabular}
\end{table}

\subsection{Text Attack}
We conducted experiments to evaluate the robustness of the GPT models against the character-level text attack. The results are summarized in Table \ref{text_attack_acc}, which shows the accuracy of each model on the original and attacked versions of the three sentiment classification datasets. The results indicate a significant drop in accuracy for all models when subjected to the character-level text attack. For instance, GPT-4o's accuracy on the IMDB dataset drops from 96.4\% to 79.3\%, highlighting its vulnerability to character-level perturbations. Similarly, GPT-4, GPT-4-turbo, and GPT-3.5-turbo exhibit substantial decreases in performance across all datasets.

All models show a notable decrease in accuracy when exposed to character-level perturbations. This suggests that even minor errors introduced by the character-level text attack can significantly impact the performance of LLMs. Among the models, GPT-4o demonstrates the highest original accuracy across all datasets, but it also experiences the most significant performance drop. GPT-3.5-turbo, while having lower original accuracy, shows a relatively smaller decrease, indicating a potential trade-off between baseline performance and robustness. The SST-2 dataset appears to be the most challenging for the models, with the highest performance drop observed for GPT-3.5-turbo (from 86.9\% to 62.2\%). This may be due to the fine-grained sentiment labels in SST-2, which make the task more sensitive to input perturbations.

\subsection{Jailbreak Attack}
\begin{table}[h]
\centering
\caption{Jailbreak Attack Accuracy for GPT Model Familiy on JailbreakHub dataset}
\begin{tabular}{c|ccc}
\hline
\multirow{2}{*}{} & \multicolumn{3}{c}{\textbf{JailbreakHub}} \\
                  & detected  & total  & percentage  \\ \hline
gpt-4o            & 1344      & 1405   & 95.7\%      \\
gpt-4             & 1263      & 1405   & 89.9\%      \\
gpt-4-turbo       & 1284      & 1405   & 91.4\%      \\
gpt-3.5-turbo     & 687       & 1405   & 48.9\%      \\ \hline
\end{tabular}
\end{table}

We conducted experiments to assess the robustness of different GPT models against jailbreak prompts using the JailbreakHub\cite{shen2024donowcharacterizingevaluating}, which includes 1405 prompts designed to bypass LLM safety mechanisms. The results, summarized in Table 1, show the number of detected jailbreak attempts and their corresponding detection percentages for each model. GPT-4o demonstrated the highest robustness, detecting 95.7\% of the jailbreak prompts. GPT-4-turbo and GPT-4 followed with detection rates of 91.4\% and 89.9\%, respectively. In contrast, GPT-3.5-turbo detected only 48.9\% of the prompts, indicating a significant vulnerability to jailbreak attacks.

The results reveal a clear difference in the effectiveness of the models in identifying and mitigating jailbreak attempts. Newer models like GPT-4o exhibit significantly higher robustness, reflecting advancements in their safety mechanisms. However, the lower detection rate of GPT-3.5-turbo underscores its susceptibility to these attacks. This emphasizes the need for ongoing enhancements in LLM development to improve their security and resilience against adversarial prompts.

\section{Conclusion}
In this paper, we evaluated the robustness of GPT family models against adversarial attacks, including character-level attacks and jailbreak prompts. Character-level attacks caused substantial accuracy drops across sentiment classification datasets, while jailbreak prompts highlighted varying levels of safety mechanism effectiveness, with GPT-4o demonstrating the highest resilience and GPT-3.5-turbo showing the greatest susceptibility. These results emphasize the need for improved adversarial training and safety mechanisms to enhance LLM robustness, ensuring reliable performance and secure deployment in critical applications. Future research should focus on developing adaptive defense strategies and robust evaluation frameworks to mitigate these vulnerabilities and foster safer deployment of LLMs in real-world contexts.

\bibliographystyle{IEEEtran}
\bibliography{IEEEabrv, bibliography.bib}

\end{document}